\documentclass{article}
\usepackage{ijcai20}

\usepackage{times}
\usepackage{soul}
\usepackage{url}
\usepackage[hidelinks]{hyperref}
\usepackage[utf8]{inputenc}
\usepackage[small]{caption}
\usepackage{graphicx}
\usepackage{amsmath}
\usepackage{amsthm}
\usepackage{booktabs}
\usepackage{algorithm}
\usepackage{algorithmic}
\usepackage{amsfonts}
\usepackage{amssymb}

\usepackage{multirow}
\usepackage{colortbl}
\usepackage{multicol}
\usepackage{fancyhdr}

\title{Stochastic Batch Augmentation with An Effective Distilled Dynamic Soft Label Regularizer}

\author{
	Qian Li$^1$$^{\ast }$\and
	Qingyuan Hu$^1$\footnote{Equal contribution}\and
	Yong Qi$^1$\footnote{Corresponding author}\and
	Saiyu Qi$^1$\and
	Jie Ma$^1$\And
	Jian Zhang$^2$\\	
	\affiliations
	$^1$School of Computer Science and Technology, Xi’an Jiaotong University, China\\
	$^2$School of Natural and Applied Sciences, Northwestern Polytechnical University, China\\
	\emails
	qianlixjtu@163.com,
	iloveavril914@stu.xjtu.edu.cn,
	qiy@xjtu.edu.cn,
	syqi@connect.ust.hk,
	\{dr.majie, jianuest\}@foxmail.com
}

\begin{document}

\maketitle

\begin{abstract}
Data augmentation have been intensively used in training deep neural network to improve the generalization, whether in original space (e.g., image space) or representation space. Although being successful, the connection between the synthesized data and the original data is largely ignored in training, without considering the distribution information that the synthesized samples are surrounding the original sample in training. Hence, the behavior of the network is not optimized for this. However, that behavior is crucially important for generalization, even in the adversarial setting, for the safety of the deep learning system.
In this work, we propose a framework called Stochastic Batch Augmentation (SBA) to address these problems. SBA stochastically decides whether to augment at iterations controlled by the batch scheduler and in which a ``distilled'' dynamic soft label regularization is introduced by incorporating the similarity in the vicinity distribution respect to raw samples. The proposed regularization provides direct supervision by the KL-Divergence between the output soft-max distributions of original and virtual data. Our experiments on CIFAR-10, CIFAR-100, and ImageNet show that SBA can improve the generalization of the neural networks and speed up the convergence of network training.
\end{abstract}

\section{Introduction}

For quite a few years, deep learning systems have persistently enabled significant improvements in many application domains, such as object recognition from vision, speech, and language and are now widely used both in research and industry.
However, these systems perform well only when evaluated on instances very similar to those from the training set. When evaluated on slightly different distributions, neural networks often provide incorrect predictions with strikingly high confidence. This is a worrying prospect since deep learning systems are increasingly being deployed in settings where the environment is noisy, subject to domain shifts, or even adversarial attacks. 

Recent research to address these issues has focused on $data$ $augmentation$ \cite{DBLP:conf/iclr/ZhangCDL18,verma2018manifold,DBLP:conf/icml/VermaLBNMLB19}, which involves producing new valid instances in input or latent space.
Although being successful, we have observed one limitation: it largely ignores the connection between the synthesized data and the original data, since they are used for training without interaction. Therefore, for the distribution information that the synthesized samples are surrounding the original sample, the model behavior is not explicitly optimized. 

To this end, we propose a dynamic soft label regularization in augmented batch to achieve this optimization that speeds up the convergence of network training, leveraging the intermediate model before final iteration and improves the generalization ability of the ultimate model. Further to reduce the redundant and high computational cost of Batch Augmentation (BA) \cite{hoffer2019augment} and allow the model to acquire higher
the capability of generalization in a short time, we induce the stochasticity on batch
augmentation inspired by the epsilon-greedy exploration in reinforcement learning.
Finally, we propose a framework dubbed Stochastic Batch Augmentation (SBA) in which batch augmentation is performed in latent space at randomly selected iterations instead of all iterations. Moreover,
so as to incorporate the prior knowledge that the derived virtual samples have a strong similarity relative to original samples, here the similarity is characterized by the KL-Divergence between their predicted distributions of neural network. 

The main contributions of the paper include the following:

\begin{itemize}
		
		\item It proposes a general framework named Stochastic Batch Augmentation (SBA), which is composed of two major ingredients: stochastic batch scheduler and distilled dynamic soft label regularization. 
		
		\item  For the first time, it proposes to employ the output distribution of reference data as the soft label to guide the neural network to fit and recognize the vicinity of the reference data. In this way, we can speed up the learning process and acquire better generalization over other methods. 
		
		\item Experimental results demonstrate that SBA can significantly improve the generalization of the neural network. In fact, it often can achieve better prediction performance due to the majority vote among the virtual and original data predictions in test time.  
	
\end{itemize}

\section{Related Work}

\begin{figure*}[t]
	\centering
	\includegraphics[width=0.80\textwidth]{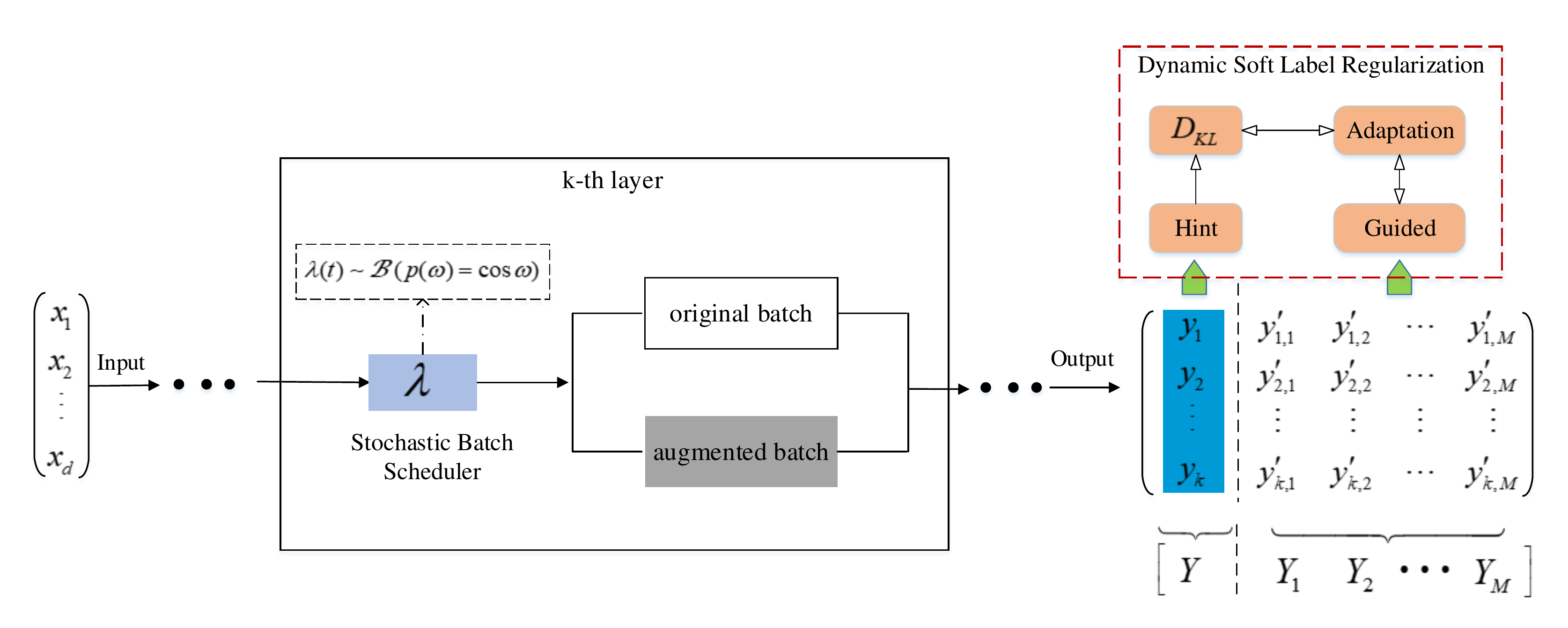}
	\caption{ Overview of our framework Stochastic Batch Augmentation. (1) Stochastic Batch Scheduler is a Bernoulli process whose observed value $\lambda(t) $ decides whether to perform augmentation at iteration $t$; (2) Distilled dynamic soft label is used to guide the model to fit the vicinity of raw sample so that model can achieve better generalization quickly.}
	\label{Fig-Overview}
\end{figure*}

A common practice in training modern neural networks is to use data augmentations – multiple instances of input samples, each with a different transformation applied to it. Common forms of augmentation include random crops, horizontal flipping, and color augmentation for image \cite{bartlett2013advances}, enhancing generalization to translation, reflection, and illumination, respectively.  
Closely related to the topic of our work is batch augmentation (BA) \cite{hoffer2019augment}: replicating instances of samples within the same batch with different data augmentations.

Data augmentations were repeatedly found to provide efficient and useful regularization, often accounting for a significant portion of the final generalization performance. One of the most common approaches of regularization includes dropout \cite{srivastava2014dropout}, and the information bottleneck \cite{wilson2017marginal}, which involves regularizing deep networks by perturbing their hidden representations. 
Another related regularization technique called Mixup was introduced by \cite{DBLP:conf/iclr/ZhangCDL18}. Mixup uses a mixed input from two separate samples with different classes and uses as target their labels mixed by the same amount.

\section{Methodology}
In this section, we present the framework named Stochastic Batch Augmentation (SBA), where the mini-batch is augmented at randomly selected epochs. The framework is illustrated in Figure \ref{Fig-Overview}. The network is equipped with a scheduler whose status is stochastically updated at the beginning of each epoch. And that scheduler decides whether to perform batch augmentation (BA) in the new epoch. During training, the intermediate output distribution of the raw sample is used to construct a supervision signal via the Kullback-Leibler divergence respect to predicted distributions of the virtual point, which takes advantage of the informative prediction of the raw point.

\subsection{Stochastic Batch Augmentation }
Suppose we have dataset $\mathcal{D}=\left \{ (\mathbf{x}_{i},y_{i}) \right \}_{i=1}^{N}$ with features $\mathbf{x}_{i}\in \mathbb{R}^{d}$ and label $y_{i}\in \left [ 1,k \right ]$
represents the label for $k$-class classification problem. Denote a classifier by $f(\mathbf{x},\boldsymbol{\theta} ):{{\mathbb{R}}^{d}}\to {{\mathbb{R}}^{k}}$, with $\boldsymbol{\theta}$ being the shared parameter. The standard training goal is to obtain the  classifier $ f $ by minimizing the average loss function $\ell(\boldsymbol{\theta} )$ over the underlying data distribution $\mathbf{P}(\mathbf{x},y)$ , following the principle of expected risk minimization:
\begin{equation}
\mathcal{L}(\boldsymbol{\theta} )
={{\mathbb{E}}_{(\mathbf{x},y)\sim \mathbf{P}(\mathbf{x},y)}}[\ell(f(\mathbf{x},\boldsymbol{\theta} ),y)]
\end{equation}
Equation (1) is often be approximated by $empirical$ $risk$ $minimization$ (ERM) on the collected samples $\mathcal{D}$
due to the distribution $\mathbf{P}(\mathbf{x},y)$ being unknown, as follow.
\begin{equation}
\mathcal{L}(\boldsymbol{\theta} )=\frac{1}{N}\sum_{i=1}^{N} \ell(f(\mathbf{x}_{i},\boldsymbol{\theta} ),y_{i})
\end{equation}

While efficient to compute, the empirical risk  monitors the behavior of $ f $ only at a finite set of $ N $ examples, causing the over-fitting and sample memorization in the 
large neural network \cite{DBLP:journals/corr/SzegedyZSBEGF13}. 
Hence, we follow the principle of $vicinity$ $risk$ $minimization$ (VRM) \cite{chapelle2001vicinal} during the training process, aiming to improve the adversarial robustness, i.e., to preserve the label consistency in small perturbation neighborhood. The VRM principle targets to minimize the $vicinity$ $risk$ $\mathcal{L}'$ on the virtual data pair $(\mathbf{x}',{y}')$ sampled from a vicinity distribution $\mathbf{P}'(\mathbf{x}',{y}'|\mathbf{x},y)$ generated from the original training set distribution $P(\mathbf{x},y)$ and consequently, the VRM-based training objective can be described as:
\begin{equation}
\mathcal{L}'(\boldsymbol{\theta} )={{\mathbb{E}}_{(\mathbf{x}',{y}')\sim \mathbf{P}'(\mathbf{x}',{y}'|\mathbf{x},y),(\mathbf{x},y)\sim \mathbf{P}(\mathbf{x},y)}}[\ell(f(\mathbf{x}',\boldsymbol{\theta} ),{y}')].
\end{equation}

However, Equation (3) still can't directly come into effect in our case, due to the batch scheduler. In general, the role of scheduler should be included in our optimization objective. Before that, we describe the workflow of the framework. First,  select a random layer $k$ from a set of eligible layers $S$ in the neural network, then process the input batch until reaching that layer. Whether to perform batch augmentation operation at layer $k$ depends on the  status $ \lambda $ of the inside scheduler,
whose status evolvement can be described by the Bernoulli discrete-time stochastic process $\mathcal{B}$ with fixed probability $ p $, 
which takes only two values, canonically $0$ and $1$ corresponding non-augmentation and augmentation respectively. Finally,  continue processing from the hidden state $k$ to the output. 

More formally, we can redefine our neural network function $y=f(\mathbf{x},\boldsymbol{\theta})$ in terms of $k:f(\mathbf{x},\boldsymbol{\theta})={{g}_{k+1}}({{h}_{k}}(\mathbf{x}))$, where ${{h}_{k}}$ denotes the mapping from an input sample to its internal DNN representation at layer $k$, and ${{g}_{k+1}}$ denotes the part mapping such hidden representation at layer $k$ to the output $y$. Let $\mathbf{X}$ denote input batch, and $\mathbf{\mathbf{X}}_{k}=h_{k}(\mathbf{X})$ be the activitions of $k$-th layer. The augmented batch of $\mathbf{X}_{k}$ is $\mathbf{X}'_{k}$. Assuming that each sample in $\mathbf{X}_{k}$ and $\mathbf{X}'_{k}$ is i.i.d, thus our optimization objective can be expressed as :
\begin{equation}
\begin{split}
\mathcal{L}''(\boldsymbol{\theta} )=&{{\mathbb{E}}_{(\mathbf{x},y)\sim \mathbf{P}(x,y),
		(\mathbf{x}'_{k},{y}'_{k})\sim \mathbf{P}'_{k}(\mathbf{x}'_{k},{y}'_{k}|\mathbf{x}_{k},y_{k})}}  \\
&{{\mathbb{E}}_{\lambda(t) \sim \mathcal{B}}} 
[\ell({g}_{k+1}(\mathbf{x}'_{k},\lambda(t) ),{y}'_{k})].
\end{split}
\end{equation}
where  $ \lambda(t) $ represents the specific status of the scheduler at the iteration $ t $. And it is worthwhile to realize that the whole process of stochastic optimization 
for Equation (4) is biased towards the vicinity distribution or original distribution determined by the probability $ p $. 

\subsection{Stochastic Batch Scheduler}

The essential of the batch scheduler is the generation of random variable $\lambda(t) $,  
which decides whether to perform augmentation at each iteration. For simplicity, we adopt Bernoulli process $\mathcal{B}$, $\lambda(t) \sim \mathcal{B}(p(\omega )=cos(\omega) )$, $\omega\in \left [ 0,\frac{\pi}{2} \right ]$, $\omega $ is used to control the skewness of that distribution. For the Bernoulli process, the probability of performing augmentation for each iteration is fixed and predefined, which is the simplest case in comparison to constructing a delicate changeable probability with respect to iteration in the way of learning rate scheduler.

The idea is motivated by the epsilon-greedy, a simple heuristics for exploration in reinforcement learning (RL). In RL, exploration, and exploitation is a well-known tradeoff. The exploitation only uses the learned policy (maybe sub-optimal) to take action at $t$-th step. In neural network training, that could correspond to the learned weights and biases up to $t$-th epoch. Instead, exploration encourages the agent to explore the faced environment. In our case, we model the data manifold at $k$-th layer as an environment with uncertainty. Thus, it is immediately obvious that the current trained neural network model is the agent, and the exploration strategy is our stochastic batch scheduler.

The stochastic scheduler makes our scheme crucially different from \cite{hoffer2019augment,verma2018manifold,shimada2019data}, in which augmentation is deterministic to be performed at all iterations, ignoring the already learned capability of generalization in neural network, i.e., the intermediate learning parameter values, for example, the updated weight after 50 epochs even if the total epochs is 200,
which can bring out higher generalization in comparison to the randomly initialized weight while lower than the ultimate converged weight values. So, they lead to a high computational cost relatively despite the performance improvement. In particular, when the selected $k$-th layer is close to the input layer, the overhead of extra computational cost is vastly obvious. Note that $\omega=\frac{\pi}{3}$,  it is a fair schedule with respect to original and augmented mini-batch, which means that $\mathbf{X}_{k}$, $\mathbf{X}'_{k}$ have the same probability of being transformed by the latter part of network $g_{k+1}(\cdot )$.

\subsection{Dynamic Soft Label Regularization}
The $k$-th layer input batch $\mathbf{X}_{k}\in {\mathbb{R}}^{B\times Q_{k}}$ is the reference of  virtual batch 
$\mathbf{X}'_{k}\in {\mathbb{R}}^{(M\times B)\times Q_{k}}$ and
$\mathbf{X}'_{k}\in {\mathbb{R}}^{(M\times B)\times Q_{k}}$ is 
sampled from the vicinity distribution $\mathbf{P}'$. $B$ is batch size, and $Q_{k}$ is width of the $k$-th layer.

In order to cause a robust cover of underlying local regime of each instance in  $\mathbf{X}_{k}\in {\mathbb{R}}^{B\times Q_{k}}$,  we combine the truncated Gaussian noise \cite{csato2001sparse} and Dropout \cite{srivastava2014dropout} to construct a mixed vicinity distribution as $\mathbf{P}'$. Given a  reference point 
$\mathbf{x}_{k}\in \mathbf{X}_{k}$, the corresponding set of virtual points can be represented by 
$ \mathbf{V}_{g} $ and $ \mathbf{V}_{d} $ as below: 
\begin{equation}
\mathbf{V}_{g}= \left \{ \mathbf{x}_{kj} \right \}_{j=1}^{p},\mathbf{V}_{d}=\left \{ \mathbf{x}_{ki} \right \}_{i=1}^{q}, 
\end{equation} 
\begin{equation}
\mathbf{X}'_{k}=\left [ \mathbf{X}_{k}; \mathbf{V}_{g};\mathbf{V}_{d}\right ], 
\end{equation}
where $p$, $q$ are the augmented number of instances respectively, $M=p+q$ is the total augmented fold. 
Specifically, the two kinds of virtual points can be
written as: 
\begin{equation}
\mathbf{x}_{kj}=\mathbf{x}_{k}+\boldsymbol{\Sigma} \times clip(\varepsilon_{kj},\tau  ),
\end{equation}
\begin{equation}
\mathbf{x}_{ki}=\mathbf{B}_{i} \odot  \mathbf{x}_{k},
\end{equation}
where $\mathbf{B}_{i}$ is a binary vector with entry $0$ denoting the deletion of a feature and $\odot$ denotes the element-wise product. The $\varepsilon_{kj} \in \mathbb{R}^{Q_{k}}$ is a vector of zero-mean independent Gaussian noise $\varepsilon_{kj} \sim \mathcal{N}(\mathbf{0} ,\sigma^{2} \mathbf{I})$ and is clipped to range $\left [ -\tau,\tau \right ]$, $\tau$ is the maximum scale for each components in generated noise. Note that we adopt a random normalized basis matrix $\boldsymbol{\Sigma} \in \mathbb{R}^{{Q_{k}}\times {Q_{k}}}$ for random direction projection,
in order to cause the diversity of derived points. And that basis matrix is regenerated when the random variable 
$ \lambda $ is set to one, aiming to avoid the unnecessary basis matrix update
in those iterations with non-augmentation.

In contrast to \cite{yang2018bamboo} that treat virtual points as an uniform distributions within $\ell_{p}$-radius ball of reference point, then use the one-hot encoding of the ground truth label $\mathbf{1}_{Y_{ref}}$ of reference point to compute the hard loss of derived virtual points, respect to the output distribution $\hat{Y}_{rel}$. 
More concretely, it can be expressed by
\begin{equation}
\ell(g_{k+1}(\mathbf{x}_{kl}))=CE(\mathbf{1}_{Y_{ref}},\hat{Y}_{rel}),
\end{equation} 
which is the standard cross entropy loss of classification for virtual point similar to reference point and $CE$  is the cross entropy function.
 
Motivated by the paradigm of knowledge distillation, i.e., teacher-student training \cite{hinton2015distilling} and adversarial training in which networks should regularly behave in the neighborhood of training data, whether in the input space or latent space.  Hence we propose to treat the soft-max output of reference point as ``soft'' label (the standard notion of soft label is the output prediction of another trained large model with high generalization) which contains more information for modeling relations between the reference and derived data, and acts as hit for the network itself as shown in Figure 1.

Different from the conventional knowledge distillation in which the soft label for a weaker student model, comes from the wiser teacher, however the soft label of our method is provided by the student itself, which also plays the role of teacher on account of the increasing generalization during the training phase as indicated by the learning curve. In this sense, we can view the updating learning parameters as the immature recognition ability of classifier resulting from the previous training iterations.

\begin{algorithm}[t]	
	\caption{Training procedure of SBA }
	\label{Alg-Clustering}
	\begin{algorithmic}[1]
		\FOR   {$t$ $\leftarrow$ $1$ to $n_{train\underline{\hspace{0.3em}}steps}$} 
		\STATE {Sample from Bernoulli process $\mathcal{B}$, $\lambda(t) \sim \mathcal{B}(p(\omega ))$}  \\
		\STATE {Update parameters by primary objective -$ \frac{\partial \mathcal{I}_{1}}{\partial \boldsymbol{\theta}}$} \\
		\IF    {$\lambda(t)$=1}
		\STATE {Update parameters by the conservative constraint -$\frac{\partial \mathcal{I}_{2}}{\partial \boldsymbol{\theta}}$} 
		\ENDIF
		\ENDFOR
	\end{algorithmic}
\end{algorithm}

Exploiting it to guide the model to learn to fit and recognize the virtual points more quickly, thus leading towards faster convergence. And by doing so, the eventually trained model is to be more insensitive to adversarial perturbations.
Note that the soft label is dynamically changed during training since learning parameters are continuously updated with the iterations. 
In another aspect, it essentially incorporates explicit similarity between the reference and virtual points in the local vicinity, where the relevant notion of similarity is based on the output distribution of the neural network. Thus, the so-called soft loss of virtual points can be represented by:
\begin{equation}
\ell(g_{k+1}(\mathbf{x}_{kl}))=D_{KL}(\hat{Y}_{ref}\parallel \hat{Y}_{rel}),
\end{equation}

where $D_{KL}$ is Kullback–Leibler divergence, a measure of how one probability distribution is different from another reference probability distribution. 
Note that this method can be regarded as a conservative constraint on the predicted distributions of neural network model in the local area of each point in the transformed space.

\subsection{Learning and Inference}
Training is to minimize the Equation (4).
Because of the induced run-time stochasticity on mini-batch
augmentation, the training iteration objective and parameters update differ a lot in form and meaning, which depending on the observed value of $ \lambda(t) $ at 
iteration $ t $. With the help of indicator function $\mathbb I$
, the optimization problem at iteration $t$ can be written as follows:
\begin{equation}
\begin{split}
\mathcal{L}''_{t}(\boldsymbol{\theta} )=&\frac{1}{B}\sum_{i=1}^{B}CE(\mathbf{1}_{Y_{i}},Y_{i})+ \\
&\frac{\eta \cdot  \mathbb I(\lambda(t))}{M\cdot   B}\sum _{i=1}^{B}\sum _{j=1}^{M}D_{KL}(Y_{i}\parallel Y_{ij}),
\end{split}
\end{equation}
where $\eta$ is the constant to balance the two terms in Equation (11) which is easy to be solved by any SGD algorithm. $Y_{i}$ is the output distribution of reference point and $Y_{ij}$ is its corresponding output distribution of virtual point. The equation consists of two components: the first part $\mathcal{I}_{1}$ minimizes the objective function on raw mini-batch, aiming to achieve full utilization of the mini-batch with more weight. The second component $\mathcal{I}_{2}$ is a similarity constraint which limits the dramatic changes of predictions in the local vicinity of reference point, which is dropped when no augmentation performed (i.e., $ \lambda(t) = 0 $). 

\begin{figure}[!t]
	\centering
	\includegraphics[width=0.32\textwidth]{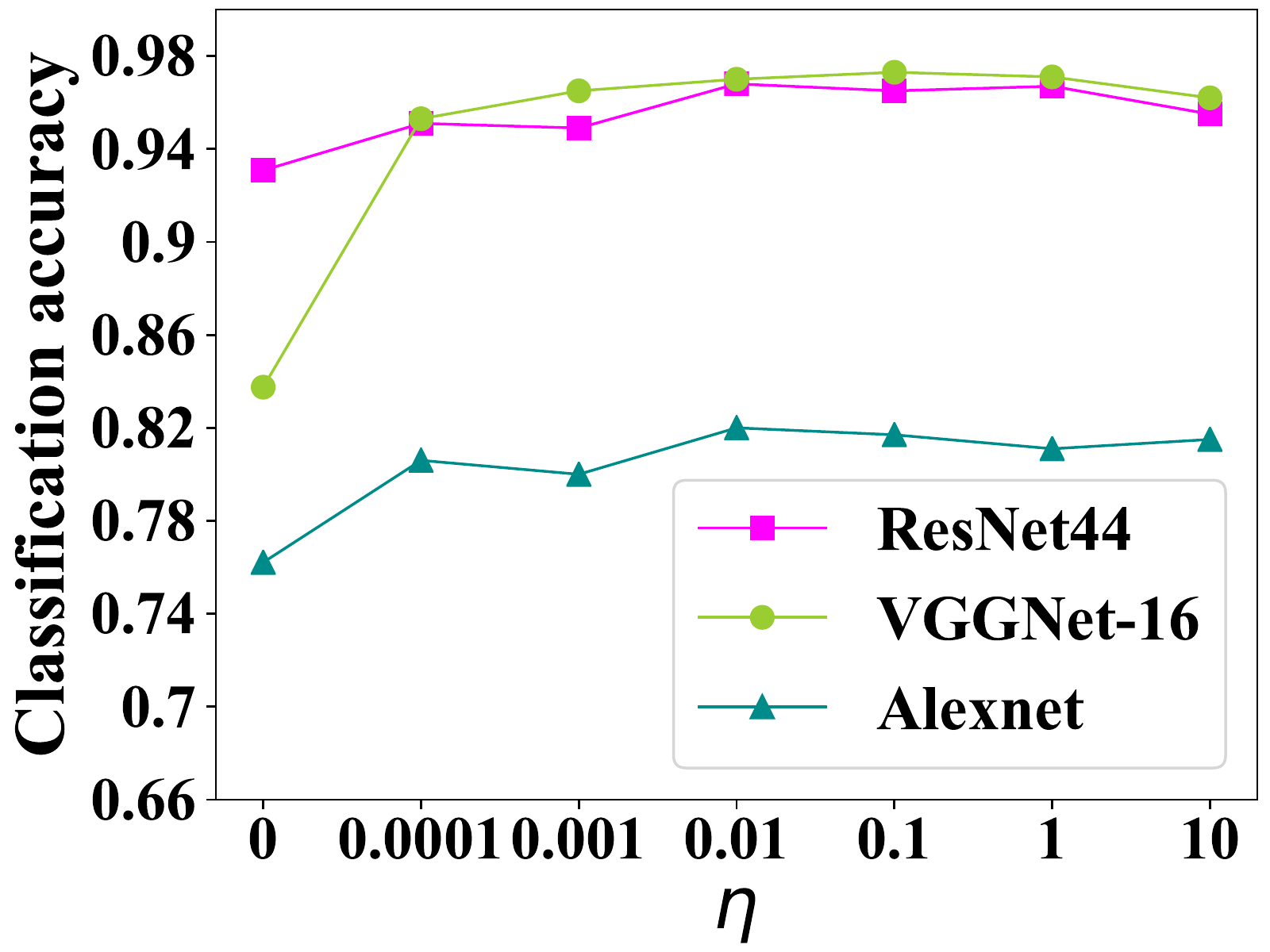}  \\
	\includegraphics[width=0.32\textwidth]{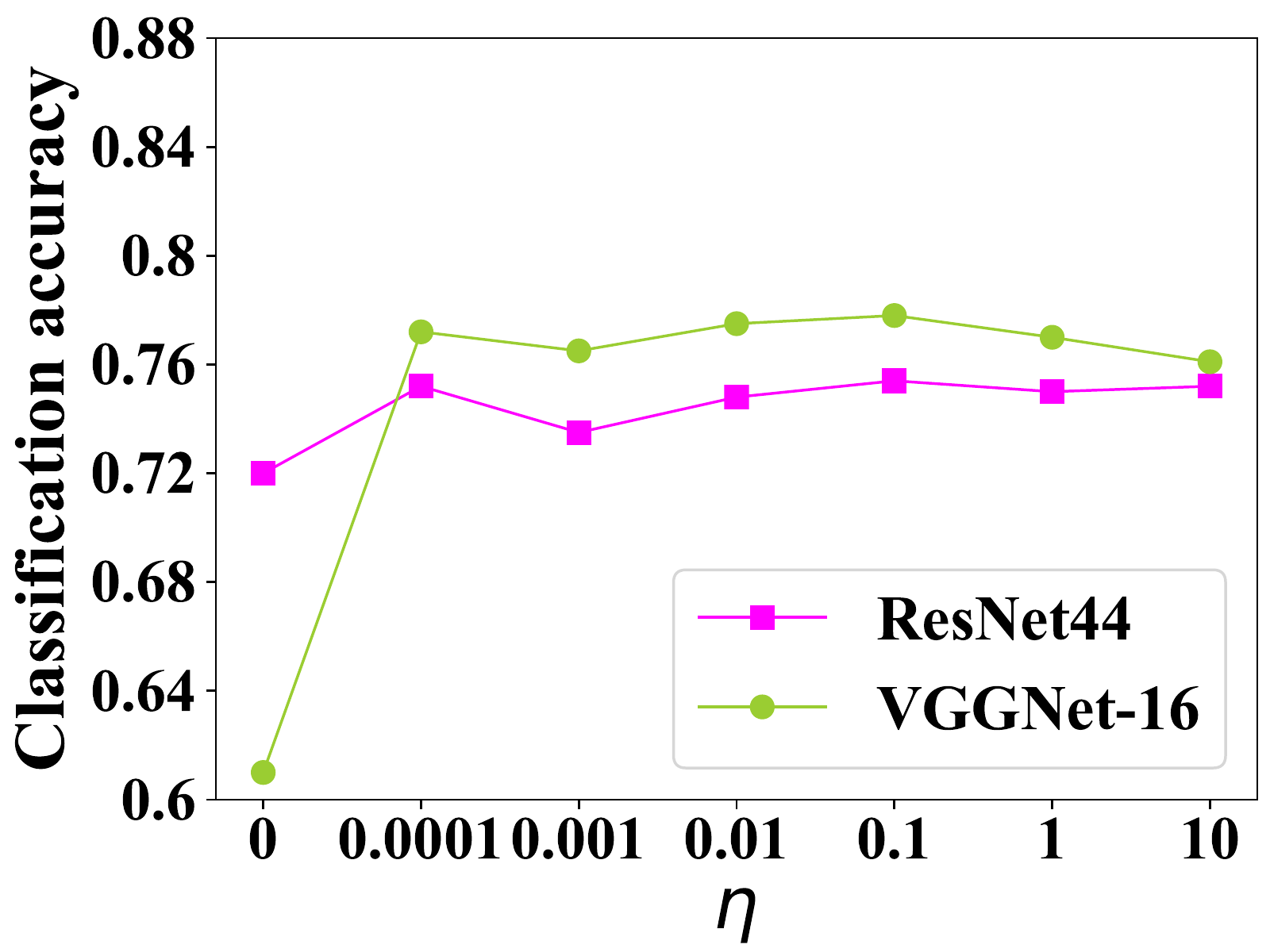}
	\caption{ Performance result under different parameter $\eta$ on CIFAR-10 (above) and CIFAR-100 (below).}
	\label{Fig-Hyper-parameters}
\end{figure}

\begin{figure*}[t]
	\centering
	\includegraphics[width=0.33\textwidth]{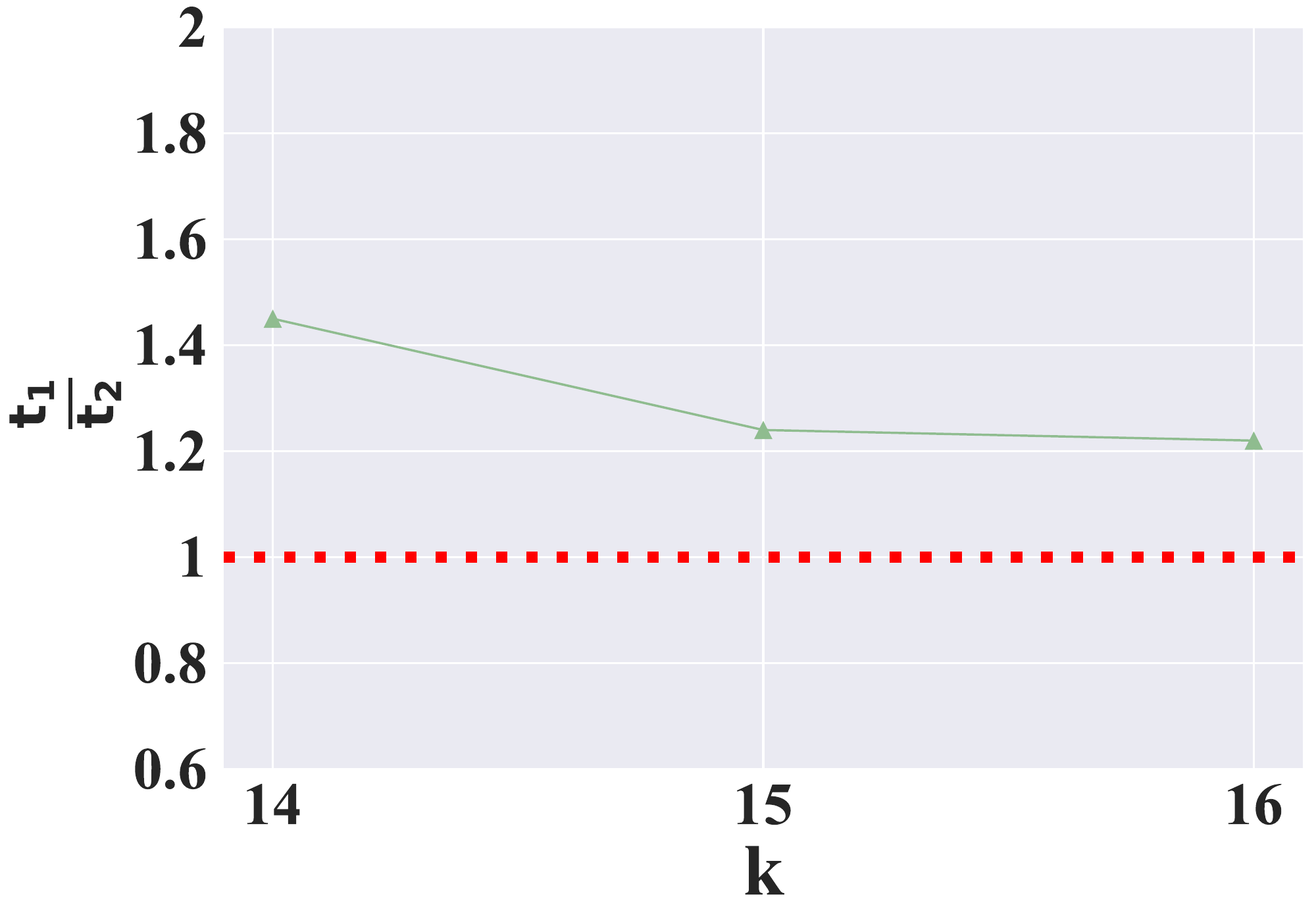} 
	\includegraphics[width=0.33\textwidth]{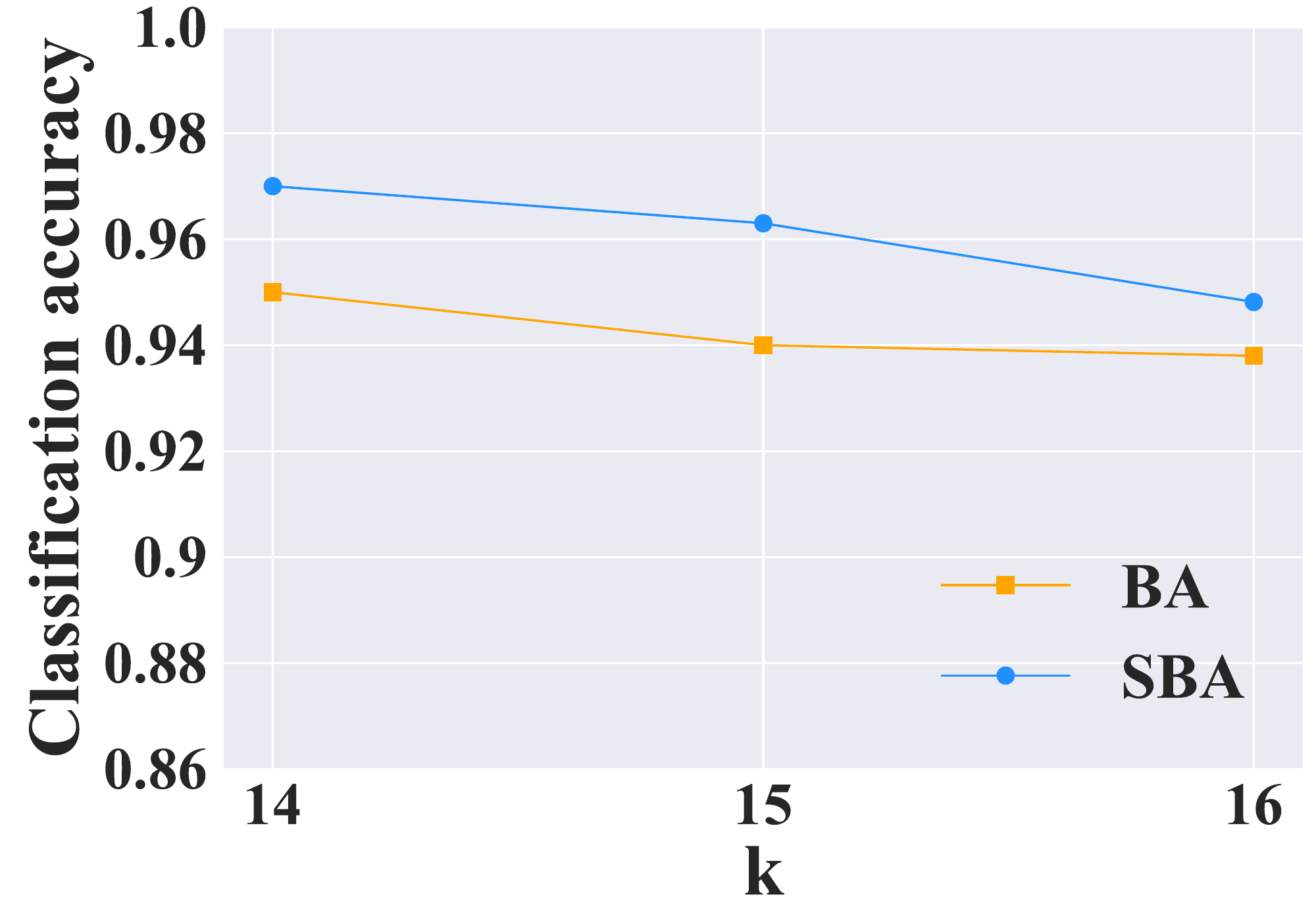}  \\      
	\includegraphics[width=0.33\textwidth]{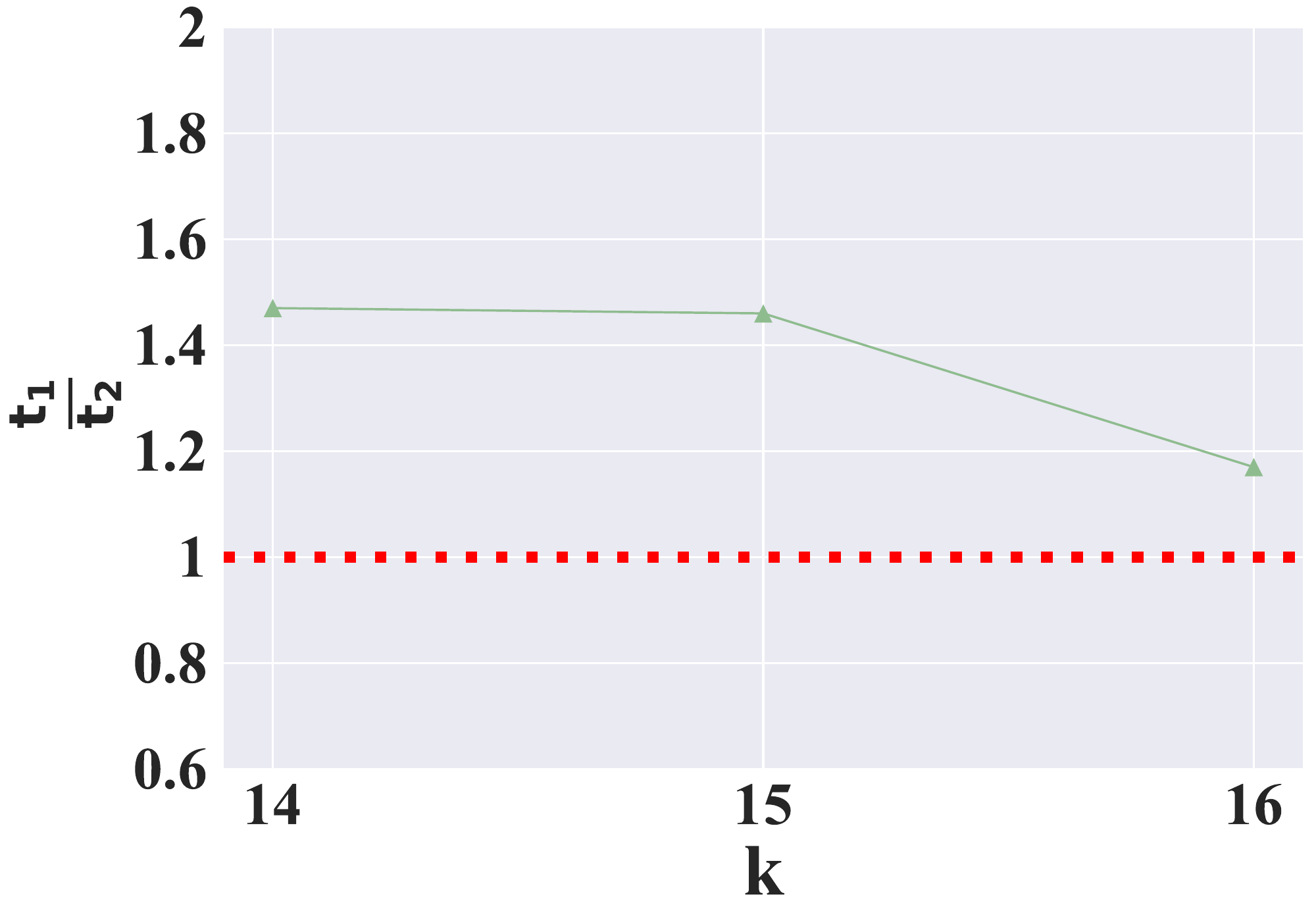} 
	\includegraphics[width=0.33\textwidth]{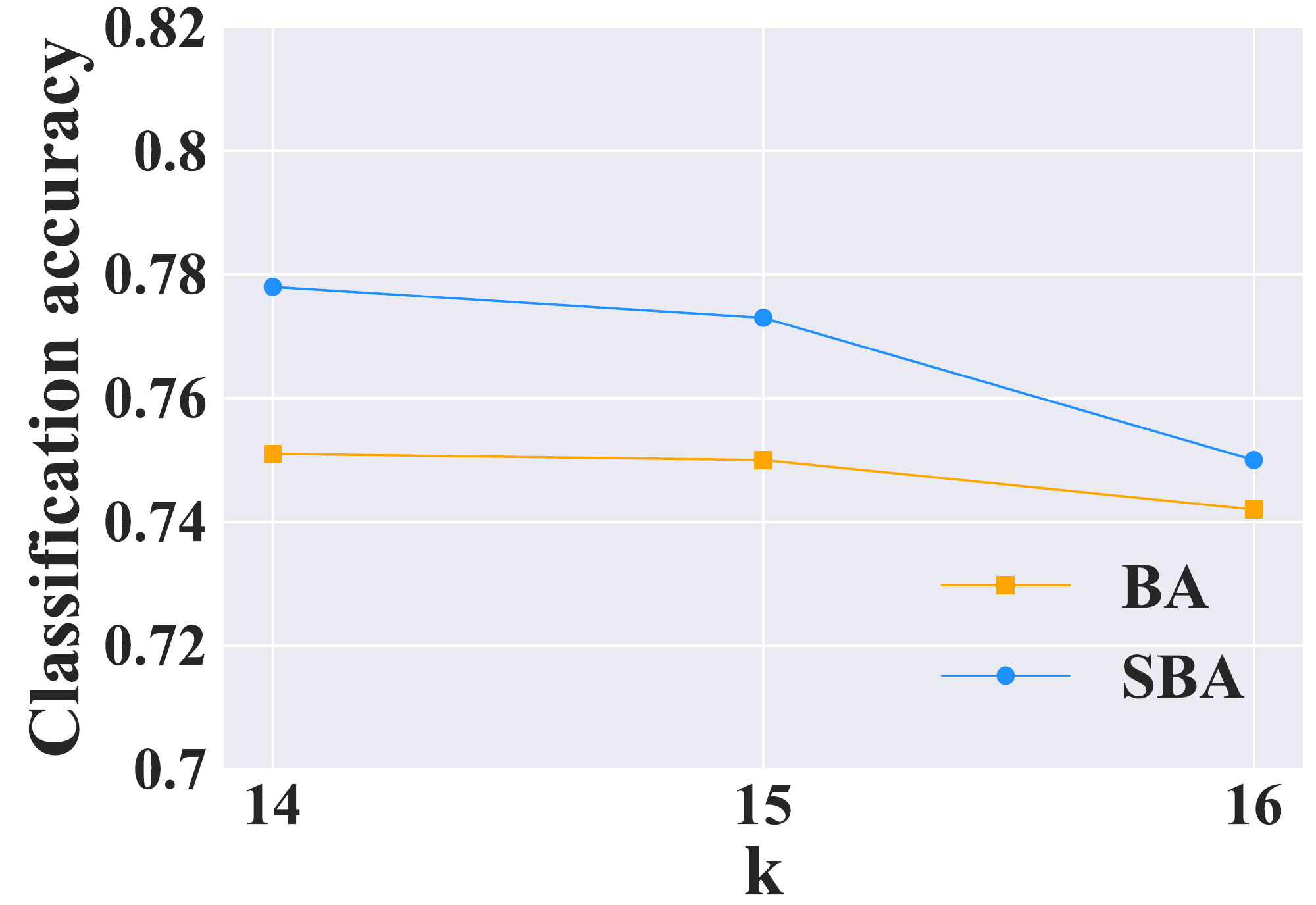}
	\caption{ Performance comparison for SBA and BA on the CIFAR-10 (above) and CIFAR-100 (below) datasets using VGGNet-16. X-axis: $\mathbf{k}$-th; Y-axis: Calculated cost ratio $\frac{t_{1}}{t_{2}}$ (left),  Test accuracy (right).}
	\label{Fig-time-acc}
\end{figure*}

To understand the effect of second term further, consider the gradient of it with respect to the learning parameters $ \boldsymbol{\theta} $:
\begin{equation}
\frac{\partial \mathcal{I}_{2}}{\partial \boldsymbol{\theta}}=\frac{\mathbb I(\lambda(t))}{M\cdot   B}\sum _{i=1}^{B} \sum_{j=1}^{M}\frac{\partial }{\partial \boldsymbol{\theta}} Y_{i}^{k}(\log Y_{i}^{k}-\log Y_{ij}^{k})
\end{equation}
where the upper script $ k $ indexes the component of vector and the small $\varepsilon _{i}^{k}$ is the difference between the $Y_{i}^{k}$ and $Y_{ij}^{k}$ which must be nearly identical, conforming the principle of adversarial training as mentioned in previous section. Then  utilize the linear approximation of $\log Y_{ij}^{k}$:
\begin{equation}
\begin{split}
\log Y_{ij}^{k}&=\log (Y_{i}^{k}+\varepsilon _{ij}^{k}) \\
&\approx \log Y_{i}^{k}+\frac{\varepsilon _{ij}^{k}}{Y_{i}^{k}\ln 2}
\end{split}
\end{equation}
the Equation (12) then can be simplified as:
\begin{equation}
\begin{split}
\frac{\partial \mathcal{I}_{2}}{\partial \boldsymbol{\theta}}&\approx \frac{ \mathbb I(\lambda(t))}{M\cdot   B}\sum_{i=1}^{B}\sum_{j=1}^{M}\left ( -\frac{1}{\ln 2}\cdot \frac{\partial \varepsilon _{ij}^{k}}{\partial \boldsymbol{\theta}} \right )  \\
&=\frac{ \mathbb I(\lambda(t))}{B}\sum_{i=1}^{B}\frac{\partial \boldsymbol \Delta _{i}}{\partial \boldsymbol{\theta}}
\end{split}
\end{equation}
In Equation (14), the inner term $\boldsymbol \Delta _{i}=-\frac{1}{M\ln 2}\sum_{j=1}^{M}\varepsilon _{ij}^{k}$ depicts the inconsistency of the network's behaviour in the neighbourhood of sample $i$, so in this sense, the $\mathcal{I}_{2}=\frac{\mathbb I(\lambda(t))}{B}\sum_{i=1}^{B} \boldsymbol \Delta_{i}$ is the average inconsistency over the raw mini-batch.  
And the supervisory signal $\frac{\partial \mathcal{I}_{2}}{\partial \boldsymbol{\theta}}$ leverages the complementary 
information in the predicted distribution \cite{DBLP:conf/iclr/ChenWLCPCWJ19} to minize the inconsistency, varied from the first primary objective $\mathcal{I}_{1}$ wherein solely 
exploits the information from the ground-truth class for supervision  
as follows:
\begin{equation}
\frac{\partial \mathcal{I}_{1}}{\partial \boldsymbol{\theta}}=\frac{1}{B}\sum_{i=1}^{B}\frac{\partial \log Y_{i}^{c_{i}}}{\partial \boldsymbol{\theta}}
\end{equation}
where the upper script $c_{i}$ denotes the true class of sample $ i $. The Algorithm $1$ describe the training mechanism.

In inference stage, we use the simple majority vote strategy to decide the final outcome of test sample.

\begin{table*}[!t]
	\centering
	\newcommand{\tabincell}[2]{\begin{tabular}{@{}#1@{}}#2\end{tabular}}
	\scalebox{0.85}{
		\begin{tabular}
			{p{1.7cm}<{\centering} p{2.7cm}<{\centering}
				p{1.7cm}<{\centering}p{1.7cm}<{\centering}
				p{3.2cm}<{\centering}p{3.2cm}<{\centering}p{1.7cm}<{\centering}}
			\toprule[1.5pt]
			Dataset & Network  &  Baseline & 	Cutout & Cutout + BA & SBA & Relative Impro.    \\
			\midrule[1pt]
			\multirow{3}*{CIFAR-10} & ResNet44  & 6.92\%  & 6.28\%  & 4.56\% ($M=40$) & 3.23\% ($M=40$) & \textbf{53.32\%} 	 \\
			~                      & VGGNet-16  & 16.25\%  & 6.17\% & 4.69\% ($M=32$) & 2.73\% ($M=32$) & \textbf{83.20\%}    \\ 
			~                      & AlexNet    & 23.88\%  & 21.34\% & 20.00\% ($M=32$)  & 18.03\% ($M=32$)      &     \textbf{24.50\%}  \\ \hline
			\\
			\multirow{2}*{CIFAR-100} & ResNet44  & 28.01\% & 27.04\% & 25.85\% ($M=40$) & 24.61\% ($M=40$)    
			&  \textbf{12.14\%}   \\
			~                        & VGGNet-16 & 38.67\% & 27.00\% & 24.69\% ($M=32$) & 22.19\% ($M=32$)    &   \textbf{42.62\%}      \\ 
			\hline \\
			\multirow{3}*{ImageNet} & ResNet50  & 23.73\% &  $\times$ & 23.16\% ($M=4$) & 22.72\% ($M=4$)    
			&  \textbf{4.26\%}   \\
			~                        & VGGNet-16 & 23.78\% &  $\times$  & 21.29\% ($M=4$) & 19.37\% ($M=4$)    &   \textbf{18.54\%}      \\ 
			~                        & AlexNet   & 41.69\% & $\times$ & 37.73\% ($M=8$)  & 34.40\% ($M=8$)       &   \textbf{17.49\%}    \\   
			\bottomrule[1.5pt]
		\end{tabular}	
	}
	\caption{Error rates obtained by the testing methods on CIFAR-10, CIFAR-100 and ImageNet: Relative improvement in error/ppl over baseline is listed in percentage.}
	\label{Tab-Errorrates}
\end{table*}

\section{Experiments}

\subsection{Experimental Setup}
Our experiments encompass a range of datasets CIFAR-10, CIFAR-100 \cite{krizhevsky2009learning}, ImageNet \cite{russakovsky2015imagenet}, and different kinds of network architectures VGGNet-16 \cite{DBLP:journals/corr/SimonyanZ14a}, AlexNet \cite{krizhevsky2012imagenet}, ResNet44, and ResNet50. We compare the original training baseline, as well as the following methods: Cutout \cite{devries2017improved}, and Batch Augmentation (BA) \cite{hoffer2019augment}.

Throughout our experiments, for each of the models, unless explicitly stated,
we tested our approach using the original training regime and data augmentation described by its authors. To support our claim, we did not change the learning rate used or the number of epochs.
To reduce the unnecessary hyper-parameter tuning and make strong baselines for different datasets and networks, we adopt the reported best hyper-parameters for $ M $ in \cite{hoffer2019augment}, while it has shown that the batch augmentation can improve the models' performance consistently for a wide range of augmentation fold. 
As for the regularization parameter ${\eta }$ , we optimize through a simple grid-search procedure on held-out validation data over the ranges $\eta\in \left \{0, 0.0001,0.001,0.01,0.1,1,10 \right \} $. For all the results, the reported performance value (accuracy or error rate) is the median of the performance values obtained in the final 5 epochs.

\subsection{Impact of Hyper-parameters}
As mentioned above, Stochastic Batch Augmentation (SBA) has one important hyper-parameter: $\eta$. We first implement a group of experiments to quantify the effect that the parameter has on model classification accuracy. More specifically, we allow the parameter $\eta$ to range from $0$ to $10$ with tenfold increments, both at training and testing time. 

We present results on SBA using the Alexnet, VGGNet-16, and ResNet44 architecture, corresponding to different parameter $\eta$. These results for CIFAR-10 and CIFAR-100 are in Figure \ref{Fig-Hyper-parameters}.
The Alexnet and ResNet44 models achieve the highest test accuracy when parameter $\eta$ is $0.01$, and the VGGNet-16 models achieve the highest test accuracy when parameter $\eta$ is $0.1$ on CIFAR-10. The VGGNet-16 and ResNet44 models achieve the highest test accuracy when parameter $\eta$ is $0.1$ on CIFAR-100. Therefore, we adopt $0.01$ and $0.1$ as parameter $\eta$ expectation for experiments that follow, as a constant to balance the two terms in Equation (11).
We can observe that the hyper-parameter $\eta$ can perform well in the range $\left [ 0,10 \right ]$, without significant efficiency on the generalization of networks.

\subsection{Efficiency of SBA} 
Data Augmentation is a general method, so the SBA and  BA can be added to any set of the latent transformed feature while the metric matters for the latent space. In a manner similar to Dropout, our experiments typically apply the augmentation to fully connected layers towards the deep end of the network due to features in the last few layers, which are more likely to lie in the Euclidean space.
To evaluate and compare the efficiency of representations learned with SBA and BA, we train VGGNet-16 models on the CIFAR-10 and CIFAR-100. 

Figure \ref{Fig-time-acc} shows the experimental results. $t_{1}$ and $t_{2}$ represent the runtime required to achieve convergence (the accuracy is almost stable) for BA and SBA, respectively. Validation convergence speed of SBA has noticeably improved compared to BA (left), with a significant increase in final validation classification accuracy (right). 
We can observe that SBA can decrease the redundant computation. SBA not only can faster convergence be achieved with stochastic batch scheduler that can eliminate unnecessary computation within each iteration, but increases accuracy as well. Note that the test accuracy of BA and SBA both decrease. This may be due to the improper metric in the corresponding transformed space (e.g., induced noise is undesirable). However, the SBA can consistently outperform BA. 

Moreover, we managed to achieve high validation accuracy much quicker with SBA. We trained a ResNet44 with SBA on CIFAR-10 for two-thirds of the iterations needed for the BA, using a larger learning rate and faster learning rate decay schedule. 
This indicates not only an accuracy gain but a potential runtime improvement for given hardware. We can infer that the main reason is that traditional batch augmentation is deterministic to be performed at each iteration per epoch, ignoring the already learned capability of generalization in the neural network. So, they lead to high computational cost relatively. In particular, when the selected $\mathbf{k}$-th layer is close to the input layer, the overhead of extra computational cost is more obvious.

\subsection{Performance of SBA }
To show the effectiveness of our method, we empirically investigate the performance of SBA on three datasets: CIFAR-10, CIFAR-100, and ImageNet. Our experiment compares SBA to the baseline model and other state-of-art methods. 
As can be seen from Table \ref{Tab-Errorrates}, the relative error reduction of SBA over the baseline is at least 4.26\%, and with a large margin in some cases. For example, on the CIFAR-10 dataset, the relative error reduction achieved by SBA is more than 24\%.  
VGGNet-16 trained with SBA achieves an error rate of 2.73\% on CIFAR-10, which is even 13.52\% better than the baseline. 
Notice that this gain is much larger than the previous gains obtained by Cutout + BA against baseline (+2.23\%), and by Cutout against baseline (+3.44\%).

Our proposed SBA achieves an error rate of 34.40\% with AlexNet on ImageNet, which outperforms the Cutout + BA by more than 3.33\%. Overall, our results justify that SBA can improve the generalization of the neural network, which originates from the explicit data augmentation on the latent space as well as the conservative constraint on the predicted distributions for virtual samples.

\section{Conclusion}
In this paper, we have presented a framework named Stochastic Batch Augmentation (SBA) for improving the generalization of deep neural network. 
In SBA, we significantly reduce the computational cost by randomized batch augmentation scheme. We also introduce a distilled dynamic soft label regularization technique for learning, which explicitly incorporates the conservative constraint on the predicted distributions. 
The experimental results on three standard datasets using standard network architectures show the superiority of our proposed framework.

\section*{Acknowledgments}

This work is supported by National Natural Science Foundation of China under Grant No.61672421.

\bibliographystyle{named}
\bibliography{ijcai20}

\end{document}